\documentclass[conference]{IEEEtran}
\IEEEoverridecommandlockouts
\usepackage{amsmath,amssymb,amsfonts}
\usepackage{algorithmic}
\usepackage{graphicx}
\usepackage{textcomp}

\usepackage{graphicx}
\usepackage{booktabs}
\usepackage{microtype}
\usepackage{xcolor}
\def\BibTeX{{\rm B\kern-.05em{\sc i\kern-.025em b}\kern-.08em
    T\kern-.1667em\lower.7ex\hbox{E}\kern-.125emX}}
\begin{document}

\title{Self-Supervised and Semi-Supervised Polyp Segmentation using Synthetic Data}


\author{%
Enric Moreu*, Eric Arazo*, Kevin McGuinness, Noel E. O'Connor \\
  Insight SFI Centre for Data Analytics, Dublin City University (DCU), Dublin, Ireland \\
  \texttt{\{enric.moreu, eric.arazo\}}{@insight-centre.org} 
}

\maketitle

\begin{abstract}
Early detection of colorectal polyps is of utmost importance for their treatment and for colorectal cancer prevention. Computer vision techniques have the potential to aid professionals in the diagnosis stage, where colonoscopies are manually carried out to examine the entirety of the patient's colon. The main challenge in medical imaging is the lack of data, and a further challenge specific to polyp segmentation approaches is the difficulty of manually labeling the available data: the annotation process for segmentation tasks is very time-consuming. While most recent approaches address the data availability challenge with sophisticated techniques to better exploit the available labeled data, few of them explore the self-supervised or semi-supervised paradigm, where the amount of labeling required is greatly reduced. To address both challenges, we leverage synthetic data and propose an end-to-end model for polyp segmentation that integrates real and synthetic data to artificially increase the size of the datasets and aid the training when unlabeled samples are available. Concretely, our model, Pl-CUT-Seg, transforms synthetic images with an image-to-image translation module and combines the resulting images with real images to train a segmentation model, where we use model predictions as pseudo-labels to better leverage unlabeled samples. Additionally, we propose PL-CUT-Seg+, an improved version of the model that incorporates targeted regularization to address the domain gap between real and synthetic images. The models are evaluated on standard benchmarks for polyp segmentation and reach state-of-the-art results in the self- and semi-supervised setups.
\end{abstract}

\section{Introduction}
\label{sec:intro}
Early colorectal cancer (CRC) detection is crucial for a successful treatment: once polyps in the colon are detected and diagnosed, early treatment leads to an increased survival rate~\cite{freeman2013early} and even the prevention of CRC development~\cite{freeman2013long}. Colonoscopies, however, are a complex screening process that require experienced medical doctors to visually explore the entirety of the patient's colon for the detection of tissue growths in colon walls, i.e.\ polyps. The recent success of computer vision techniques in medical images~\cite{pomponiu2016deepmole, oktay2018attention}, makes them a potential candidate to aid clinicians during this process and through the task of semantic segmentation provide accurate information about the location and shape of polyps. While several works already explore the application of semantic segmentation techniques for colorectal polyp detection~\cite{2021_ArXiv_hardnetmseg, zhang2020adaptive}, they all rely on fully annotated datasets and ignore the main limitation of obtaining labeled samples. For the semantic segmentation task, in particular, this is a laborious process that requires the annotators, medical professionals in this case, to label every pixel in an image.   

\begin{figure}[t]
\centering
\includegraphics[width=\linewidth]{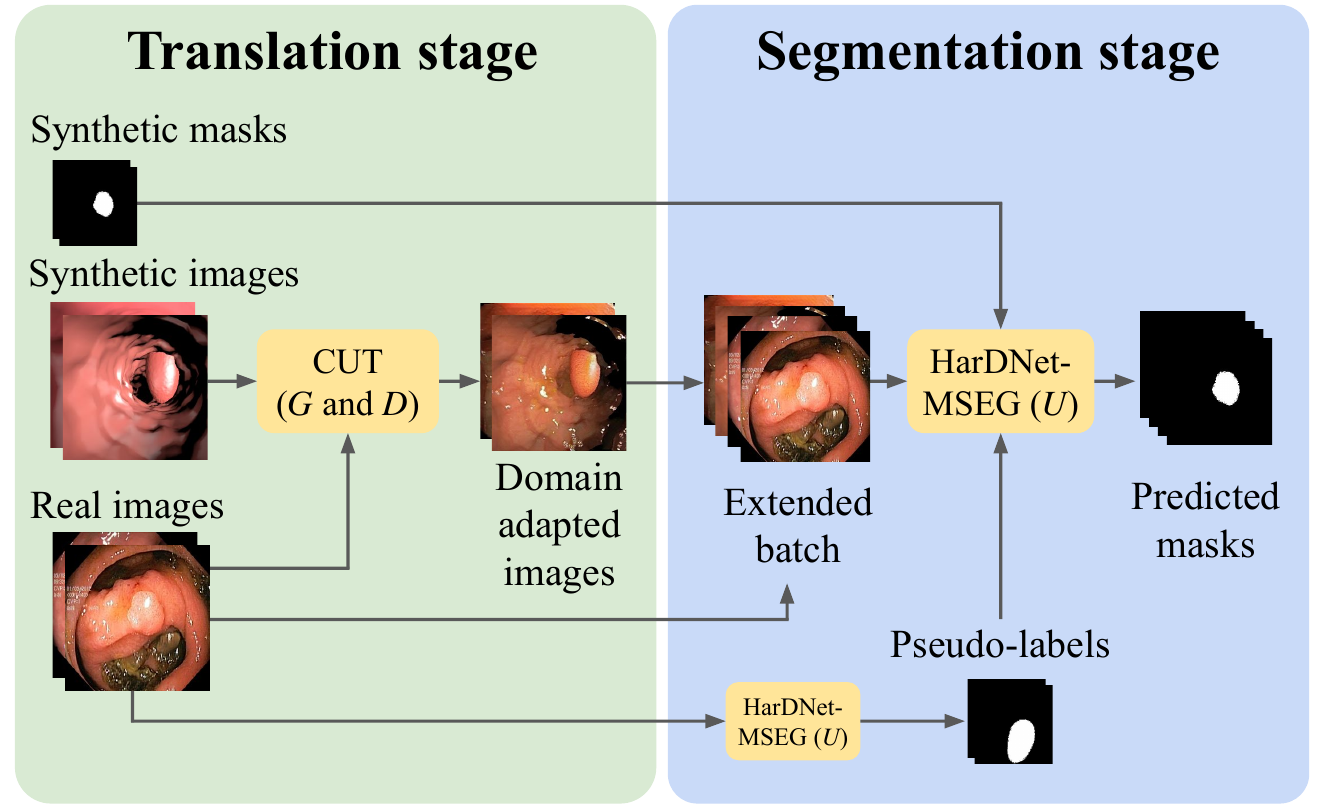}
\par
\caption{Schematic of PL-CUT-Seg in the self-supervised scenario: two-stage architecture for polyp segmentation that leverages real-world data without annotations. Our end-to-end approach utilizes synthetic images, which are translated to the real-world domain, and incorporates pseudo-labeled real-world images within each mini-batch to narrow the gap between the training and testing distributions. This forces the segmentation model, $U$, to learn features that better generalize to real-world images.~\label{fig:basic_scheme}}
\end{figure}

The reliance of current polyp segmentation approaches on carefully annotated datasets for semantic segmentation results in models trained on relatively small amounts of data, which hinders their scalability and generalization to new data. Recent works explore self-supervised and semi-supervised techniques to address this limitation~\cite{2022_ExpSysJ_joint, 2021_AICS_synthColon, 2021_ICCV_collaborative}. However, few explore the introduction of synthetic data~\cite{2022_ExpSysJ_joint, 2021_AICS_synthColon}, and, to our knowledge, none explore the combination of synthetic data with semi-supervised learning. The main limitation of the application of synthetic data is the generation of a 3D model from which to generate the images, which is often very laborious, but the specific constraints of medical imaging allows for the utilization of models that simulate the structure of a particular area of the body: the interior of the colon in this case. While synthetic data has already been shown to be useful in the self-supervised setup~\cite{2022_ExpSysJ_joint, 2021_AICS_synthColon}, its application to the semi-supervised setup is yet to be explored. This setup allows the segmentation model to be exposed to both real images and real segmentation masks during training, which narrows the gap between the training and testing distributions. Initial steps to narrow this gap have shown image translation models to be a valuable tool~\cite{2017_ICCV_CycleGAN, 2020_ECCV_cut} that can adapt the synthetic artificially-looking images to more realistic-looking ones. Image translation models adapt the textures and patterns of a source image to look like the ones in a target image while maintaining the spatial information of the source images.

We propose PL-CUT-Seg, an end-to-end approach to polyp segmentation that learns to adapt synthetic images to the realistic domain and generate segmentation masks that locate the polyps in an image. In particular, we address the main weaknesses of \cite{2022_ExpSysJ_joint} by narrowing the distribution gap between the training domain, i.e.\ synthetic images, and the testing domain, i.e.\ colonoscopy images. Moreu et al.~\cite{2022_ExpSysJ_joint} proposed a model composed of an image translation stage and a segmentation stage: while the former was exposed to both the synthetic and colonoscopy images, the latter only trained with the translated images and the synthetically generated segmentation masks. Conversely, as shown in Figure~\ref{fig:basic_scheme}, we expose the segmentation stage to both the translated images with the synthetic masks and the real images with model predictions as masks, i.e.\ pseudo-labels. Additionally, we propose an enhanced version of the model, PL-CUT-Seg+, and introduce an interpolation-based regularization technique that aims to reduce the domain gap between the real and the synthetic images, which results in improved polyp detection accuracy. Inspired by previous work on pseudo-label approaches~\cite{2020_NeurIPS_fixmatch}, we design a confidence mask that indicates which regions of the mask to be considered, i.e.\ the most confident predictions.

This paper explores the application of synthetic data to polyp segmentation under different levels of label availability. We propose to reduce the domain gap between real and synthetic data with the introduction of pseudo-labels, a careful design of the training batches, and a regularization technique that directly targets the domain gap. This approach improves the integration of synthetic data for polyp segmentation in self-supervised and semi-supervised learning, which highlights the potential for exploiting synthetic data as a solution to the data scarcity challenge in medical imaging. As a result, the proposed model PL-CUT-Seg reaches state-of-the-art performance in most of the semi-supervised and self-supervised setups in the standard datasets used in polyp segmentation: Kvasir, CVC-ClinicDB, ETIS, CVC-ColonDB, CVC-300. Additionally, we adopt the semi-supervised learning setup in polyp segmentation as a valuable benchmark for evaluating the generalization of the approaches when the data availability is reduced. Our contributions can be summarized as follows:
\begin{itemize}
    \item An end-to-end model for integrating synthetic and real data under different levels of supervision: semi- and self-supervised learning.
    \item A novel approach that integrates pseudo-labels, confidence masks, and mixup in a unified framework to address the main challenge when training with real and synthetic images.
    \item A thorough analysis of the components of the model and their effect on generalization across the available datasets for polyp segmentation.
\end{itemize}

\section{Related work}
We categorize the relevant literature into three main groups: approaches to polyp segmentation, representation learning under label scarcity, and the applicability of synthetic data in training neural networks. This section outlines the most relevant works in the context of leveraging synthetic data for polyp segmentation under label scarcity scenarios.

\subsection{Polyp segmentation}

Early approaches to polyp segmentation addressed the task from a low-level feature perspective and leveraged shapes and textures in a hand-crafted manner; Rahim et al.~\cite{2020_PubMed_Survey} provide a broad overview of the evolution of computer-aided medical imaging. Recent approaches leverage neural networks to directly learn from the data. The U-Net architecture~\cite{2015_Springer_unet}, for instance, has shown great success in medical tasks and has become the backbone for several works on polyp segmentation~\cite{sun2019colorectal, yeung2021focus} and other medical image analysis tasks~\cite{ibtehaz2020multiresunet, gaal2020attention}. This has led to improved architectures like UNet++~\cite{2018_Springer_unetPlusPlus}, which incorporates a modified skip pathway to reduce the semantic gap between encoder and decoder, and ResUNET++~\cite{2019_IEEE_resunetPlusPlus}, which adapts the ResUNET~\cite{2018_IEEE_resunet} architecture initially proposed for road segmentation. More recently, Fan et al.\ proposed PraNet~\cite{2020_Springer_pranet} to divide the computation of segmentation masks into two parallel branches, one computing a coarse mask and another introducing the finer details. HarDNet-MSEG~\cite{2021_ArXiv_hardnetmseg} is a combination of backbone and decoder architecture that, aside from producing competitive results, is very fast at inference time. Dong et al.\ proposed Polyp-PVT~\cite{2021_ArXiv_polypPVT}, a model that uses transformers~\cite{vaswani2017attention} as a more powerful feature extractor than CNN-based encoders, and introduced a number of modules to specifically address the elusive nature of polyps. Similarly, UACANet~\cite{2021_ACM_uacanet} computes saliency maps at different stages of the network and introduces the uncertainty of feature maps to inform model predictions.

All currently existing approaches address data scarcity in different ways, but few of them explore the degradation of the model's performance when the availability of training data is reduced. In particular, the semi-supervised and self-supervised learning scenarios are only addressed in~\cite{2022_ExpSysJ_joint, 2021_AICS_synthColon, 2021_ICCV_collaborative}.

\subsection{Scarce annotations}
Recent research has drastically improved the capacity of neural networks to learn when the labeled data is scarce (semi-supervised learning) or unavailable (self-supervised learning). Semi-supervised learning approaches can be divided into two categories: consistency regularization and pseudo-labeling approaches. The former leverage the unlabeled data by encouraging the model to produce similar outputs for a single image under different perturbations~\cite{2017_NeurIPS_meanTeacher, 2019_NeurIPS_mixmatch, 2022_Elseiver_ICT}; the latter use the predictions of the model as labels for the unlabeled samples~\cite{2020_IJCNN_PL, 2019_CVPR_labelProp, 2021_AAAI_curriculumPL}. Current state-of-the-art methods combine both alternatives in a holistic approach to semi-supervised learning~\cite{2020_NeurIPS_fixmatch, 2021_NeurIPS_flexmatch, 2021_ICCV_PAWS}. Self-supervised learning, on the other hand, initially addressed the lack of labels by using proxy tasks as a supervisory signal, e.g.\ predicting image rotations~\cite{2018_ICLR_rotations}, inferring color~\cite{2016_ECCV_colorization}, or predicting relative position of image patches~\cite{2015_ICCV_context}. Later methods, however, show a shift towards instance-based learning: these guide the training by encouraging features from patches of one image to be closer together while pushing away the features from patches from different images~\cite{2021_CVPR_simsiam, 2021_ICLR_imix, 2020_CVPR_MoCo}. 

Advances in the semi-supervised learning approaches have permeated the semantic segmentation task~\cite{2021_ICCV_segmSSL, 2022_CVPR_PLSegm, 2022_ICLR_bootstrapping} resulting in methods that propagate the information from a few annotated masks to the rest of the unlabeled samples. Despite the existence of some works applying these techniques to medical imaging~\cite{2019_Springer_medImgSSL, 2020_CVPR_focalmix, 2020_ArXiv_roam}, few have explored the polyp segmentation task under label scarcity. To the authors' knowledge, only one work explores this paradigm~\cite{2021_ICCV_collaborative}, which proposes an approach that can leverage unlabeled data alongside labeled data and investigates the behaviors of this approach with different levels of labeled samples.

\subsection{Synthetic data for polyp segmentation}
Current medical imaging research addresses the data scarcity problem by designing approaches that better leverage the data available, but few explore the degradation in performance when the training labeled data availability is reduced. In the face of this challenge, some works already leverage synthetic data in a variety of domains~\cite{2018_CVPR_learning, 2018_ECCV_synthUrban, 2019_CVPR_SegmSynth, 2016_CVPR_synthia}. Similarly, in the medical domain, several works propose techniques to leverage this type of data~\cite{2018_ISBI_synthDAGAN, 2021_ArXiv_improving, 2021_TMI_lowAvailability, 2017_ArXiv_dualGAN}. More related to this work, Moreu et al.~\cite{2021_AICS_synthColon} focus on polyp segmentation and propose a training strategy to combine synthetic images with real ones in a self-supervised setup, which removes the middle step of annotating the colonoscopy images. They also release a synthetic dataset for polyp segmentation generated from a 3D rendered model and later improve the approach by end-to-end training~\cite{2022_ExpSysJ_joint}. This approach leverages image-to-image translation models~\cite{2020_ECCV_cut, 2017_ICCV_CycleGAN} to adapt the synthetic images and reduce the gap between the training and testing distributions. In this work, we go a step further and address the domain gap between the adapted images and the images at test time: we leverage the few available real samples (labeled or not) and, through the use of pseudo-labels, expose the segmentation network to combinations of real and synthetic-adapted images and the corresponding masks.    


\section{Method}
In the context of semantic segmentation and given a dataset $\mathcal{D}$ composed of pairs $(x_i, y_i)$ of images and segmentation masks, a model $U$ estimates the segmentation mask $y_i \in [0,1]^{H\times W \times K}$, where $K=1$ is the number of classes. The segmentation mask is obtained by predicting a label for each pixel in the input image $x_i \in \mathbb R^{H\times W\times 3}$. For the polyp segmentation task, this translates into a binary problem where each pixel is labeled as colon or as polyp: $y_i^{(n,m)} = 1$ if the pixel $x_i^{(n,m)}$ corresponds to a polyp and $y_i^{(n,m)} = 0$ otherwise (background).

Available datasets for colorectal polyp segmentation are fully labeled, and each image $x_i$ has a ground truth mask $y_i$ associated. As a result, semi-supervised and self-supervised learning for this domain are under-explored, and the techniques to exploit additional unlabeled data are underdeveloped. Since the main characteristic of these setups is the lack of ground truth labels, i.e. segmentation masks, we define $\mathcal{D}_l$ as the set of labeled samples where each sample $x_i$ has an associated mask $y_i$ and $\mathcal{D}_u$ as the set of unlabeled samples where the corresponding masks are unavailable.

\begin{figure*}[t]
\centering
\includegraphics[width=15cm]{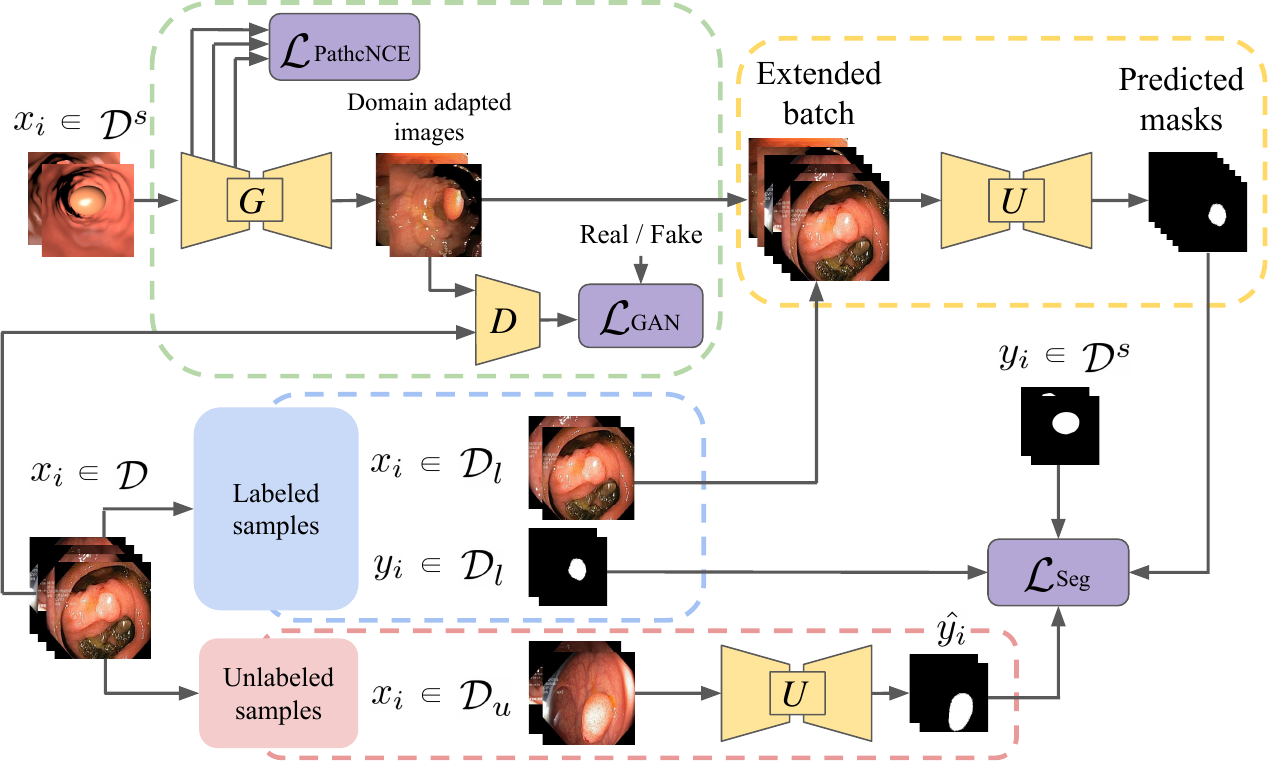}
\par
\caption{Self- and semi-supervised model architecture that utilizes a combination of adapted synthetic images, real-world images with annotations, and real-world images with pseudo-labels in each mini-batch. This approach exposes the model to both real-world data and synthetic data and enables it to learn features that are better suited for real colonoscopy images. The blue and red dashed lines encapsulate the labeled and unlabeled samples respectively, the green dashed lines encapsulate the domain adaptation stage, and the orange ones the segmentation stage.~\label{fig:complete_scheme}}
\end{figure*}

In particular, we leverage an additional set $\mathcal{D}^s$ of synthetic image-masks pairs $(x_i, y_i)$ that are computer-generated and do not require colonoscopies or annotation, which are laborious and time-consuming processes. The main challenge when using this data is the domain disparity between synthetic and real images. To address this we propose PL-CUT-Seg, a model that leverages the predictions from the segmentation network as pseudo-labels along with the available ground truth masks $y_i \in \mathcal{D}_l$. This model is inspired by previous work by Moreu et al.~\cite{2022_ExpSysJ_joint}, CUT-Seg, which consists of an unpaired image-to-image translation stage~\cite{2020_ECCV_cut} that addresses the image-to-image translation part of the problem and a segmentation stage that generate the masks from adapted synthetic images. By introducing pseudo-labels in the pipeline, PL-CUT-Seg is able to better integrate the unlabeled samples from the real set $\mathcal{D}_u$ with the synthetic data in $\mathcal{D}^s$. Figure~\ref{fig:complete_scheme} provides an overview of the model structure.

\subsection{Image-to-image translation stage}
The first stage of PL-CUT-Seg translates synthetic images to the real domain based on images of the available colorectal polyp segmentation datasets. This part of the model is based on CUT~\cite{2020_ECCV_cut}, a generative model for unpaired image-to-image translation composed of a generator $G$ and a discriminator $D$ that are trained in an adversarial fashion~\cite{2014_NeurIPS_gans} to generate real-looking images $z_i$ from synthetic images: $G(x_i) = z_i$. Additionally, two contrastive learning terms in the loss encourage the model to maintain spatial consistency in both input images: synthetic and real. Since we seek to leverage the synthetically generated masks, we give a higher weight to the term corresponding to the spatial consistency of synthetic images. As detailed in \cite{2020_ECCV_cut, 2022_ExpSysJ_joint}, the CUT loss function is given by
\begin{multline}
    \mathcal{L}_{\text{GAN}}(G, D, x_s, x_r) + \lambda_{x_s}\mathcal{L}_{\text{PatchNCE}}(G, H, x_s) +\\
    \lambda_{x_r}\mathcal{L}_{\text{PatchNCE}}(G, H, x_r),
\label{eq:cut_loss}
\end{multline}
where $x_s$ and $x_r$ are the synthetic and real images, $\mathcal{L}_{\text{GAN}}(G, D, x_s, x_r)$ corresponds to the adversarial loss~\cite{2014_NeurIPS_gans}, and $\mathcal{L}_{\text{PatchNCE}}(G, H, x_s)$ and $\mathcal{L}_{\text{PatchNCE}}(G, H, x_r)$ are the contrastive terms that encourage spatial consistency between the generated images and the input images for the synthetic and real domains, respectively. These terms minimize a contrastive term from a two-layer perceptron $\mathit{H}$ that projects input patches to a normalized feature space. Then the hyperparameters $\lambda_{x_r}$ and $\lambda_{x_s}$ control the contribution of each of these terms.

\begin{figure}[t]
\centering
\includegraphics[width=0.15\textwidth]{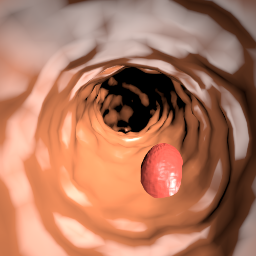}
\includegraphics[width=0.15\textwidth]{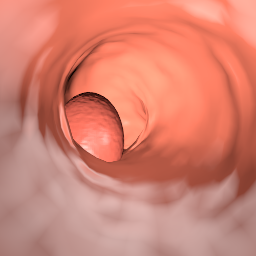}
\includegraphics[width=0.15\textwidth]{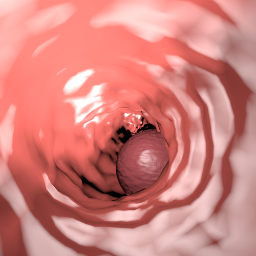}

\includegraphics[width=0.15\textwidth]{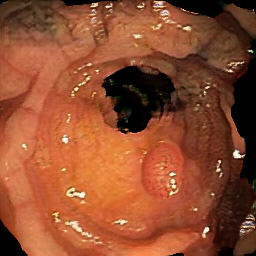}
\includegraphics[width=0.15\textwidth]{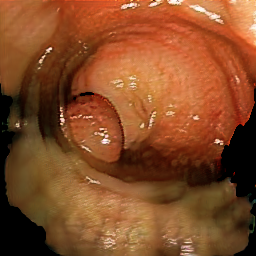}
\includegraphics[width=0.15\textwidth]{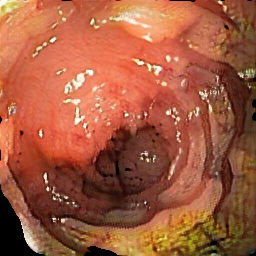}

\caption{Synthetic images (first row) converted to the real-world domain (second row).} \label{converted_polyps}
\end{figure}

This image-to-image translation stage directly addresses the domain gap between real and synthetic images by adapting the style of the synthetic images to a more realistic one. These adapted images, however, are still far from images obtained from a colonoscopy. This hinders the applicability of segmentation models trained on these images to real colonoscopy images (see Figure~\ref{converted_polyps}). Hence, it is essential to address the remaining domain gap between the generated images $z_i$ and the realistic images $x_r$. We address this in the segmentation stage of PL-CUT-Seg.

\subsection{Polyp segmentation stage}
The segmentation stage of PL-CUT-Seg consists of a segmentation network $U$ that generates the binary masks $\hat{y}_i$ that highlight polyps in colonoscopy images $x_i$. Unlike common approaches to polyp segmentation, PL-CUT-Seg trains on combinations of real labeled and unlabelled images, and synthetic images. Inspired by CUT-Seg~\cite{2022_ExpSysJ_joint}, the transformed synthetic images $z_i$ are used to train $U$ and reduce the need for manual annotation of the real images obtained from colonoscopies: ${U}(z_i) = \hat{y_i}$. We address the domain gap between the transformed images $z_i$ and the real images in two ways: by introducing pseudo-labels and by exposing the segmentation network $U$ to real images and through additional regularization in the form of interpolation training. 

The introduction of pseudo-labels $\hat{y}_i$ as annotations for the unlabeled samples $x_i \in \mathcal{D}_u$ is the main contributor to the improved performance of PL-CUT-Seg. We use model predictions as segmentation masks and combine the unlabeled samples with the labeled real samples and the synthetic samples in the segmentation stage. The pseudo-labels are initially defined as all-background masks and are updated with model predictions $p_i = U(x_i)$ where $x_i \in \mathcal{D}_{u}$ through an additional forward pass at the end of every training epoch. The loss function corresponding to the segmentation stage is
\begin{equation}
    \mathcal{L}_\text{Seg} = \sum_{x_i \in \mathcal{D}}\mathcal{L}_\text{DICE}(y_i, p_i),
\label{eq:dice_loss}
\end{equation}
where $\mathcal{D} = \mathcal{D}_l \cup \mathcal{D}_u \cup \mathcal{D}^S$, $p_i$ are the model prediction for $x_i$, and $y_i$ correspond to the available label for the labeled samples, i.e. $x_i \in \mathcal{D}_l$ or $x_i \in \mathcal{D}^S$, and to the pseudo-label $\hat{y}_i$ for the unlabeled, i.e. $x_i \in \mathcal{D}_u$. Then $\mathcal{L}_\text{DICE}$ is the DICE loss widely used in semantic segmentation~\cite{sanchez2020deep}.

We address the domain gap between real and synthetic samples by using samples from both domains as inputs for the segmentation stage and create each batch $B$ with half of the samples from the real domain and half of them from the synthetic domain: $B = \{s_1, \cdots, s_{\frac{M}{2}}, t_1, \cdots, t_{\frac{M}{2}}\}$, where $s_i \in \mathcal{D}_l \cup \mathcal{D}_u$, $t_i \in \mathcal{D}^S$, and $M$ is the batch size. In the case of self-supervised learning, all real images are associated with a pseudo-label $\hat{y}_i$, and in the case of semi-supervised learning half of the real samples come from the labeled subset $\mathcal{D}_l$ with the corresponding manual annotations and the other half from the unlabeled set $\mathcal{D}_u$ with the corresponding pseudo-label. 

Additionally, for improved results, we propose PL-CUT-Seg+, which incorporates interpolation training as a regularization technique to reduce the domain gap and a pseudo-labeling masking step to reduce the effect of incorrect predictions. Interpolation-based techniques have shown to be an effective solution in semi-supervised learning to reduce the gap between labeled and unlabeled distributions in the training set~\cite{2022_Elseiver_ICT} and to avoid overfitting the labeled samples~\cite{2020_IJCNN_PL}. The interpolation strategy is inspired by MixUp data augmentation~\cite{2018_ICLR_mixup}, where the model is trained on convex combinations of samples and the corresponding labels, and by previous research boosting semi-supervised learning with interpolation training~\cite{2020_IJCNN_PL, 2020_NeurIPS_fixmatch}. Concretely, we extend the batch $B$ with the random interpolation of the images in $B$, which results in $\{s_1, \cdots, s_{\frac{M}{2}}, t_1, \cdots, t_{\frac{M}{2}}, i_{1}, \cdots, i_{M}\}$, where $i_n = \lambda s_p + (1-\lambda) s_p$, $\lambda$ is randomly drawn from a beta distribution as introduced in~\cite{2018_ICLR_mixup} (see Section \ref{section:experiments} for details on the parameters used in the beta distribution), and $s_p$ and $s_q$ are randomly selected from $B$. The masks are interpolated following the same strategy. Unlike~\cite{2020_TMI_threshold}, which applied interpolation training in polyp segmentation, we use soft segmentation masks where each pixel is the interpolated value of the two corresponding masks. This avoids the need to set up a threshold to obtain binary masks after the interpolation.

Finally, following recent methods on semi-supervised learning~\cite{2020_NeurIPS_fixmatch, 2021_NeurIPS_flexmatch}, we avoid guiding the model training with incorrect predictions by using only the most confident predictions. This is often implemented as a filtering step where samples associated with lower confidence model outputs are discarded. We adapt this to the semantic segmentation setup and mask out the pixels with lower confidence during the computation of the loss. As a result, the DICE loss only considers the pixels of the unlabeled samples that are associated with higher confidence values. Concretely, PL-CUT-Seg+ only considers the pixels associated with predictions with over 0.999 confidence, for the background and polyps.

\subsection{Training PL-CUT-Seg}
End-to-end trained models tend to outperform those that are trained in several stages~\cite{2019_ICCV_DeepVCP, 2017_arxiv_deal, 2021_CVPR_soho}. This is often attributed to better cooperation between features from different layers and additional freedom from the end-to-end model to learn better-suited features for the end task. As demonstrated in~\cite{2022_ExpSysJ_joint}, this also applies to colorectal polyp segmentation and synthetic data. Hence, we integrate the two stages of our PL-CUT-Seg into a single backpropagation loop and train them end-to-end.

The overall end-to-end model loss function integrates both the image-to-image translation and the segmentation loss terms. Then it trains in an adversarial two-step fashion: on the first step, a binary loss function is employed to optimize the discriminator $D$ from the CUT model to distinguish between real images (from $\mathcal{D}$ or $\mathcal{D}^S$) and images generated by the generator ($z$); and on the second step, this same binary loss is used to optimize the generator $G$ to generate images that would be classified as real by the discriminator. In this last step, the segmentation term, $\mathcal{L}_\text{Seg}$, is also included in the optimization step, as well as the contrastive terms in equation~\eqref{eq:cut_loss}, $\mathcal{L}_{\text{PatchNCE}}(G, H, x_s)$ and $\mathcal{L}_{\text{PatchNCE}}(G, H, x_r)$. Given the design of the approach, the translation loss from equation~\eqref{eq:cut_loss} only updates the parameters from $G$ and $D$, while the segmentation loss from equation~\eqref{eq:dice_loss} updates both the parameters in the segmentation network $U$ and in the discriminator and generator $G$ and $D$ from the translation stage. Note that we weigh the segmentation loss with a hyperparameter $\lambda_{\mathit{S}}$ to control the influence that the segmentation loss has on the overall training of the model.

\section{Experiments}\label{section:experiments}
This section describes the proposed semi- and self-supervised paradigms, the design of the different ablation studies, and the hyperparameters used through all the experiments. We provide a thorough analysis of the different elements involved in PL-CUT-Seg and PL-CUT-Seg+ and their role in training the model, explore the robustness to different levels of label availability, and compare the model performance to state-of-the-art approaches. 

\subsection{Setup}
The experiments in this section explore different levels of supervision by varying the percentage of labeled real samples (size of $\mathcal{D}_l$ with respect to $\mathcal{D}_u$) in the dataset. For the self-supervised setup, we assume that all the labels in the real dataset are unavailable ($\mathcal{D}_l = 0$) and train the model with $\mathcal{D}_u$. For the semi-supervised setup, we define $\beta$ as the percentage of labeled data in the training set: $\frac{\mathcal{D}_l}{\mathcal{D}_l+\mathcal{D}_u} \times 100$. Following the work in~\cite{2021_ICCV_collaborative}, we explore $\beta = 15\%$ and $\beta = 30\%$. 

The most widely adopted benchmark for colorectal polyp segmentation consists of a combination of five datasets for the testing stage and a larger set of non-overlapping samples from two of these datasets for the training stage~\cite{2020_Springer_pranet, 2021_ACM_uacanet, 2021_ArXiv_polypPVT, 2022_ExpSysJ_joint}. The testing sets consist of the following datasets: CVC-300~\cite{2017_JHE_CVC300} with 60 samples, CVC-ClinicDB~\cite{2015_Elseiver_CVCClinicDB} with 62 samples, CVC-ColonDB~\cite{2012_PR_CVCColonDB} with 380 samples, ETIS~\cite{2014_IJCARS_ETIS} with 196 samples, and Kvasir~\cite{2020_ICMM_kvasir} with 100 samples. The training set is constituted of 1450 samples: 550 samples from CVC-ClinicDB and 900 from Kvasir. For the bulk of the experiments, we split the training set into two sets, one for each dataset, and evaluate on the corresponding testing sets. We evaluate the performance of the model in all the datasets as well as the effects of training with both datasets combined. The amount of synthetic samples is maintained across experiments following the procedure in~\cite{2022_ExpSysJ_joint}: 19917 samples from a 3D model. 

\subsubsection{Training}
Following Moreu et al.~\cite{2022_ExpSysJ_joint}, we train end-to-end both networks in PL-CUT-Seg: the CUT~\cite{2020_ECCV_cut} model followed by the HarDNet-MSEG~\cite{2021_ArXiv_hardnetmseg} segmentation network initialized with ImageNet-trained weights on the encoder. We use the Adam optimizer with the default momentum values $\beta_1 = 0.9$ and $\beta_2 = 0.999$, a constant value for the learning rate of $1\times10^{-5}$ for 300 epochs. Each batch comprises 2 real images and 2 fake images in the translation stage as an input for the CUT model $G$ and $D$, and 2 translated images and 2 real images in the input of the segmentation model $U$. When training in the semi-supervised setup, half of the real images are from the labeled set $\mathcal{D}_l$ and half the unlabeled set $\mathcal{D}_u$. The batch size at the input of the segmentation model doubles when applying mixup since we keep the non-interpolated samples. We found the optimal value of the mixup $\alpha$ parameter to change for different datasets and use 2.0 for Kvasir, 0.5 for CVC-ClinicDB, and 0.5 when training with both datasets. In all the cases we maintain the data augmentation and preprocessing techniques applied in~\cite{2022_ExpSysJ_joint}: the 320$\times$320 input images are randomly cropped to 256$\times$256, horizontally and vertically flip 50\% of the times, and randomly rotated between -360 and 360 degrees. All samples are normalized with a (0.5, 0.5, 0.5) per channel mean and standard deviation. Finally, we use the default hyperparamters suggested in CUT~\cite{2020_ECCV_cut} and set $\lambda_{x_r} = 1$ and $\lambda_{x_s} = 1$ in Eq.~\eqref{eq:cut_loss}.

\subsection{Ablation study}
We conduct an ablation study to investigate the effect of the techniques employed in PL-CUT-Seg+. Table~\ref{tab:ablation1} compares CUT-Seg with the approach proposed in this paper and evidences the effect of each of the elements introduced. Experiments are conducted on the Kvasir dataset for self-supervised and semi-supervised setups with 0\%, 15\%, and 30\% labeled data. In particular, Table~\ref{tab:ablation1} shows that the utilization of pseudo-labels (``+ Pseudo-labels") greatly improves the performance of the framework, particularly by providing the model with exposure to real-world images. Furthermore, the use of Mixup and confidence masks results in further enhancement of the performance achieved with pseudo-labels. The last two rows in Table~\ref{tab:ablation1}, correspond to PL-CUT-Seg and  PL-CUT-Seg+, which show improved performance across the different levels of labeled samples. We additionally train the segmentation model on synthetic images without the image translation stage for the self-supervised setup (0\% labeled samples) and obtained an mDICE and IoU scores of 56.78 and 48.22. These results highlight the importance of the adaptation stage of the proposed model.
\begin{table}
\caption{\label{tab:ablation1} Results for Kvasir validation set for the self-supervised setup and the semi-supervised setup with 30\% of labeled samples. We report best mDICE and IoU scores. Best results are in bold.}
\resizebox{0.49\textwidth}{!}{%
\begin{tabular}{lcc|cc|cc}
\toprule 
{Labeled samples:}  & \multicolumn{2}{c}{0\% } & \multicolumn{2}{c}{15\% } &  \multicolumn{2}{c}{30\%} \tabularnewline
\midrule 
{ } & mDICE & IoU & mDICE & IoU & mDICE & IoU \tabularnewline
\midrule 

{CUT-Seg~\cite{2022_ExpSysJ_joint}}  & 63.82  & 55.24 & 77.91 & 70.58 & 80.31  & 73.18 \tabularnewline
{+ Pseudo-labels}  & 77.41  & 68.54 & 84.08 & 76.61 & 85.39  & 78.32 \tabularnewline
{+ Pseudo-labels + MixUp}  & 77.72  & 68.72 & 84.32 & 77.07 & 85.87  & 78.71 \tabularnewline
{+ Pseudo-labels + MixUp + Conf. Mask}  & \textbf{78.08}  & \textbf{68.77} & \textbf{85.52} & \textbf{78.19} & \textbf{86.94}  & \textbf{79.58} \tabularnewline

\bottomrule
\end{tabular}
}
\end{table}
\subsection{Comparison with other self- and semi-supervised approaches}
This subsection presents a comparison of Pl-CUT-Seg and Pl-CUT-Seg+ with other state-of-the-art algorithms using the Kvasir and CVC-ClinicDB datasets. Table~\ref{tab:sota_individual_sets} shows that the proposed methods achieve improved performance in semi-supervised and self-supervised setups and includes HarDNet-MSEG and UACANet, two state-of-the-art approaches for the fully supervised setup (100\% labeled samples) as a reference. The performance of PL-CUT-Seg and PL-CUT-Seg+ surpasses the other approaches when trained and evaluated on the Kvasir dataset and reduces the gap between CUT-Seg and CAL on CVC-ClinicDB. This shows the potential of introducing synthetic data in the training of polyp segmentation approaches. This is especially noticeable as the number of labels decreases, where the synthetic samples seem to have a larger impact on the model training. These results, however, also point to a limitation of the approach when training and evaluating on CVC-CliniDB alone. Our hypothesis is that the synthetic dataset is more similar to Kvasir, causing PL-CUT-Seg to be less effective in adapting the synthetic images when trained with CVC-CliniDB alone. We include this in the conclusions as future work to be explored.

\begin{table}
\caption{\label{tab:sota_individual_sets} Results from training on Kvasir and CVC-ClinicDB separately and testing on individual validation sets for the self-sup and semi-supervised setups for Kvasir and CVC-ClinicDB. We report best mDICE and IoU scores. Best results are in bold. Note that the results marked with a star * come from Wu et al.~\cite{2021_ICCV_collaborative} and the semi-supervised setup might differ slightly.}
\resizebox{0.48\textwidth}{!}{%
\begin{tabular}{lc|cc|cc}
\toprule 
{ }  & \multicolumn{1}{c}{ } & \multicolumn{2}{c}{Kvasir} &  \multicolumn{2}{c}{CVC-ClinicDB} \tabularnewline
\midrule 
{ } & {Labels} & mDICE & IoU & mDICE & IoU \tabularnewline
\midrule 

{HarDNet-MSEG~\cite{2021_ArXiv_hardnetmseg}} & 100\% & 89.72  & 80.23 & \textbf{93.36}  & \textbf{88.34} \tabularnewline
{UACANet~\cite{2021_ACM_uacanet}} & 100\% & \textbf{90.74}  & \textbf{85.93} & 90.39  & 86.35 \tabularnewline

\midrule
{Hung et al.$^*$~\cite{hung2018adversarial}} & 30\% & 75.93  & 67.95 & 76.09  & 68.02 \tabularnewline
{CAL$^*$~\cite{2021_ICCV_collaborative}} & 30\% & 80.95  & 71.63 & \textbf{89.29}  & \textbf{82.57} \tabularnewline
{CUT-Seg~\cite{2022_ExpSysJ_joint}} & 30\% & 80.31  & 73.18 & 75.43  & 67.20 \tabularnewline
{PL-CUT-Seg (Ours)}  & 30\% & 85.87  & 78.71 & 83.99  & 76.92 \tabularnewline
{PL-CUT-Seg + (Ours)}  & 30\% & \textbf{86.94}  & \textbf{79.58} & 85.10  & 78.08 \tabularnewline

\midrule
{Hung et al.$^*$~\cite{hung2018adversarial}} & 15\% & 68.39  & 56.96 & 56.88  & 47.61 \tabularnewline
{CAL$^*$~\cite{2021_ICCV_collaborative}} & 15\% & 76.76  & 67.23 & \textbf{82.18}  & \textbf{74.98} \tabularnewline
{CUT-Seg~\cite{2022_ExpSysJ_joint}} & 15\% & 77.91  & 70.58 & 73.93  & 66.59 \tabularnewline
{PL-CUT-Seg (Ours)}  & 15\% & 84.32  & 77.07 & 81.48  & 74.40 \tabularnewline
{PL-CUT-Seg + (Ours)}  & 15\% & \textbf{85.52}  & \textbf{78.19} & 81.22  & 73.90 \tabularnewline

\midrule
{CUT-Seg~\cite{2022_ExpSysJ_joint}} & 0\% & 63.82  & 55.24 & 52.25  & 44.17 \tabularnewline
{PL-CUT-Seg (Ours)}  & 0\% & 77.72  & 68.72 & \textbf{64.61}  & \textbf{55.83} \tabularnewline
{PL-CUT-Seg + (Ours)}  & 0\% & \textbf{78.08}  & \textbf{68.77} & 56.84  & 46.70 \tabularnewline

\bottomrule
\end{tabular}
}
\end{table}

Additionally, the generalization ability of the proposed approach is evaluated by training on a larger set constituted by both Kvasir and CVC-ClinicDB train sets, and evaluating on the five test sets individually: Kvasir, CVC-ClinicDB, ETIS, CVC-ColonDB, and CVC-300. Table~\ref{tab:sota_multiple_sets} shows that the proposed approach surpasses CUT-Seg, the previous attempt to introduce synthetic data in the pipeline, across all the datasets. Both PL-CUT-Seg and PL-CUT-Seg+ reach competitive results across all levels of annotations and datasets while CUT-Seg fail to generalize to certain datasets, especially in the setup with 15\% of labeled samples.  We observe that the additional regularization in PL-CUT-Seg+ provides a considerable improvement in most of the cases but seems to degrade model performance when no labels are available on the dataset, the self-supervised setup. This suggests that the increased difficulty of training only with unlabeled samples might require weaker regularization techniques.

\begin{table}
\caption{\label{tab:sota_multiple_sets} Results from training on the Kvasir and CVC-ClinicDB training sets merged together and testing on individual validation sets for the self-sup and semi-supervised setups. We report best mDICE scores. Best results are in bold.}
\centering\resizebox{0.48\textwidth}{!}{%
\begin{tabular}{lc|ccccc}
\toprule 
{ }  & \multicolumn{1}{c}{Labels} & Kvasir & CVC-Cl. & ETIS & CVC-Co. & CVC-300  \tabularnewline
\midrule 

{HarDNet-MSEG~\cite{2021_ArXiv_hardnetmseg}} & 100\% & 91.20  & 93.20 & 67.70  & 73.10 & 88.70 \tabularnewline
{UACANet~\cite{2021_ACM_uacanet}} & 100\% & 91.20  & 92.60  & 76.60 & 75.10 & 91.00 \tabularnewline
\midrule
{CUT-Seg~\cite{2022_ExpSysJ_joint}} & 30\% & 64.08  & 49.33 & 24.09  & 31.61 & 36.27 \tabularnewline
{PL-CUT-Seg (Ours)}  & 30\% & \textbf{86.82}  & 75.82 & 42.35  & \textbf{67.59} & \textbf{84.04} \tabularnewline
{PL-CUT-Seg + (Ours)}  & 30\% & 86.54  & \textbf{77.97} & \textbf{46.02}  & 63.89 & 79.40 \tabularnewline

\midrule
{CUT-Seg~\cite{2022_ExpSysJ_joint}} & 15\% & 62.70  & 34.71	 & 10.31  & 12.50 & 5.88 \tabularnewline
{PL-CUT-Seg (Ours)}  & 15\% & 85.03  & 74.71 & 36.18  & \textbf{61.79} & \textbf{85.44} \tabularnewline
{PL-CUT-Seg + (Ours)}  & 15\% & \textbf{85.71}  & \textbf{75.95} & \textbf{43.68}  & 58.05 & 70.93 \tabularnewline

\midrule
{CUT-Seg~\cite{2022_ExpSysJ_joint}} & 0\% & 67.23  & 54.40 & 31.06  & 31.45 & 36.09 \tabularnewline
{PL-CUT-Seg (Ours)}  & 0\% & \textbf{78.82}  & \textbf{66.35 }& \textbf{33.05}  & \textbf{52.83} & \textbf{56.31} \tabularnewline
{PL-CUT-Seg + (Ours)}  & 0\% & 74.00  & 57.15 & 29.82  & 43.96 & 44.78 \tabularnewline
\bottomrule
\end{tabular}
}
\end{table}
%


\section{Conclusion}
This paper presents a novel framework for self- and semi-supervised polyp segmentation that achieves competitive results in scarcely labeled datasets and successfully combines synthetic and real data. The results show the potential of synthetic data in medical imaging. Experiments demonstrate that exposing the segmentation model to both real and synthetic images is crucial for achieving improved performance. This is achieved through the use of pseudo-labels, which allow the model to reduce the gap between the training and the testing distributions. Furthermore, the results indicate that synthetic data is particularly useful in scarce label setups, where the proposed approaches PL-CUT-Seg and PL-CUT-Seg+ provide improved results compared to previous approaches, especially when generalizing across the testing benchmarks. The main limitation is the adaptation of synthetic data to specific datasets, which sometimes leads to a performance drop when training on a single dataset. Hence, we encourage further research on the generation of medical data from 3D models. We also encourage the use of scarcely labeled paradigms, i.e.\ self and semi-supervised learning, as a valuable benchmark to explore the generalization and robustness of polyp segmentation approaches.


\bibliographystyle{IEEEtran}
\bibliography{conference_101719}

\begin{thebibliography}{10}
\providecommand{\url}[1]{#1}
\csname url@samestyle\endcsname
\providecommand{\newblock}{\relax}
\providecommand{\bibinfo}[2]{#2}
\providecommand{\BIBentrySTDinterwordspacing}{\spaceskip=0pt\relax}
\providecommand{\BIBentryALTinterwordstretchfactor}{4}
\providecommand{\BIBentryALTinterwordspacing}{\spaceskip=\fontdimen2\font plus
\BIBentryALTinterwordstretchfactor\fontdimen3\font minus
  \fontdimen4\font\relax}
\providecommand{\BIBforeignlanguage}[2]{{%
\expandafter\ifx\csname l@#1\endcsname\relax
\typeout{** WARNING: IEEEtran.bst: No hyphenation pattern has been}%
\typeout{** loaded for the language `#1'. Using the pattern for}%
\typeout{** the default language instead.}%
\else
\language=\csname l@#1\endcsname
\fi
#2}}
\providecommand{\BIBdecl}{\relax}
\BIBdecl

\bibitem{freeman2013early}
H.~J. Freeman, ``Early stage colon cancer,'' \emph{World journal of
  gastroenterology: WJG}, vol.~19, no.~46, p. 8468, 2013.

\bibitem{freeman2013long}
H.~J. {Freeman}, ``Long-term follow-up of patients with malignant pedunculated
  colon polyps after colonoscopic polypectomy,'' \emph{Canadian Journal of
  Gastroenterology}, vol.~27, no.~1, pp. 20--24, 2013.

\bibitem{pomponiu2016deepmole}
V.~Pomponiu, H.~Nejati, and N.-M. Cheung, ``Deepmole: Deep neural networks for
  skin mole lesion classification,'' in \emph{2016 IEEE international
  conference on image processing (ICIP)}.\hskip 1em plus 0.5em minus
  0.4em\relax IEEE, 2016, pp. 2623--2627.

\bibitem{oktay2018attention}
O.~Oktay, J.~Schlemper, L.~L. Folgoc, M.~Lee, M.~Heinrich, K.~Misawa, K.~Mori,
  S.~McDonagh, N.~Y. Hammerla, B.~Kainz \emph{et~al.}, ``Attention {U-Net}:
  Learning where to look for the pancreas,'' \emph{arXiv preprint
  arXiv:1804.03999}, 2018.

\bibitem{2021_ArXiv_hardnetmseg}
C.-H. Huang, H.-Y. Wu, and Y.-L. Lin, ``{HarDNet-MSEG}: A simple
  encoder-decoder polyp segmentation neural network that achieves over 0.9 mean
  dice and 86 fps,'' 2021.

\bibitem{zhang2020adaptive}
R.~Zhang, G.~Li, Z.~Li, S.~Cui, D.~Qian, and Y.~Yu, ``Adaptive context
  selection for polyp segmentation,'' in \emph{International Conference on
  Medical Image Computing and Computer-Assisted Intervention}.\hskip 1em plus
  0.5em minus 0.4em\relax Springer, 2020, pp. 253--262.

\bibitem{2022_ExpSysJ_joint}
E.~Moreu, E.~Arazo, K.~McGuinness, and N.~E. O'Connor, ``Joint one-sided
  synthetic unpaired image translation and segmentation for colorectal cancer
  prevention,'' \emph{Expert Systems}, p. e13137, 2022.

\bibitem{2021_AICS_synthColon}
E.~Moreu, K.~McGuinness, and N.~E. O'Connor, ``Synthetic data for unsupervised
  polyp segmentation,'' in \emph{{A}rtificial {I}ntelligence and {C}ognitive
  {S}cience ({AICS})}, 2021.

\bibitem{2021_ICCV_collaborative}
H.~Wu, G.~Chen, Z.~Wen, and J.~Qin, ``Collaborative and adversarial learning of
  focused and dispersive representations for semi-supervised polyp
  segmentation,'' in \emph{IEEE/CVF International Conference on Computer Vision
  (ICCV)}, 2021.

\bibitem{2017_ICCV_CycleGAN}
J.-Y. Zhu, T.~Park, P.~Isola, and A.~A. Efros, ``Unpaired image-to-image
  translation using cycle-consistent adversarial networks,'' in \emph{{IEEE}
  {I}nternational {C}onference on {C}omputer {V}ision ({ICCV})}, 2017.

\bibitem{2020_ECCV_cut}
T.~Park, A.~A. Efros, R.~Zhang, and J.-Y. Zhu, ``Contrastive learning for
  unpaired image-to-image translation,'' in \emph{{E}uropean {C}onference on
  {C}omputer {V}ision ({ECCV})}, 2020.

\bibitem{2020_NeurIPS_fixmatch}
K.~Sohn, D.~Berthelot, N.~Carlini, Z.~Zhang, H.~Zhang, C.~A. Raffel, E.~D.
  Cubuk, A.~Kurakin, and C.-L. Li, ``Fixmatch: Simplifying semi-supervised
  learning with consistency and confidence,'' \emph{Advances in neural
  information processing systems}, vol.~33, pp. 596--608, 2020.

\bibitem{2020_PubMed_Survey}
\BIBentryALTinterwordspacing
T.~Rahim, M.~A. Usman, and S.~Y. Shin, ``A survey on contemporary
  computer-aided tumor, polyp, and ulcer detection methods in wireless capsule
  endoscopy imaging,'' \emph{Computerized Medical Imaging and Graphics},
  vol.~85, p. 101767, Oct. 2020. [Online]. Available:
  \url{https://doi.org/10.1016/j.compmedimag.2020.101767}
\BIBentrySTDinterwordspacing

\bibitem{2015_Springer_unet}
O.~Ronneberger, P.~Fischer, and T.~Brox, ``{U-Net}: Convolutional networks for
  biomedical image segmentation,'' in \emph{International Conference on Medical
  image computing and computer-assisted intervention}.\hskip 1em plus 0.5em
  minus 0.4em\relax Springer, 2015, pp. 234--241.

\bibitem{sun2019colorectal}
X.~Sun, P.~Zhang, D.~Wang, Y.~Cao, and B.~Liu, ``Colorectal polyp segmentation
  by {U-Net} with dilation convolution,'' in \emph{2019 18th IEEE International
  Conference On Machine Learning And Applications (ICMLA)}.\hskip 1em plus
  0.5em minus 0.4em\relax IEEE, 2019, pp. 851--858.

\bibitem{yeung2021focus}
M.~Yeung, E.~Sala, C.-B. Sch{\"o}nlieb, and L.~Rundo, ``Focus {U-Net}: A novel
  dual attention-gated cnn for polyp segmentation during colonoscopy,''
  \emph{Computers in biology and medicine}, vol. 137, p. 104815, 2021.

\bibitem{ibtehaz2020multiresunet}
N.~Ibtehaz and M.~S. Rahman, ``{MultiResUNet}: Rethinking the {U-Net}
  architecture for multimodal biomedical image segmentation,'' \emph{Neural
  networks}, vol. 121, pp. 74--87, 2020.

\bibitem{gaal2020attention}
G.~Ga{\'a}l, B.~Maga, and A.~Luk{\'a}cs, ``Attention {U-Net} based adversarial
  architectures for chest x-ray lung segmentation,'' \emph{arXiv preprint
  arXiv:2003.10304}, 2020.

\bibitem{2018_Springer_unetPlusPlus}
Z.~Zhou, M.~M. Rahman~Siddiquee, N.~Tajbakhsh, and J.~Liang, ``{UNet++}: A
  nested {U-Net} architecture for medical image segmentation,'' in \emph{Deep
  learning in medical image analysis and multimodal learning for clinical
  decision support}.\hskip 1em plus 0.5em minus 0.4em\relax Springer, 2018, pp.
  3--11.

\bibitem{2019_IEEE_resunetPlusPlus}
D.~Jha, P.~H. Smedsrud, M.~A. Riegler, D.~Johansen, T.~De~Lange, P.~Halvorsen,
  and H.~D. Johansen, ``{ResUNet++}: An advanced architecture for medical image
  segmentation,'' in \emph{2019 IEEE International Symposium on Multimedia
  (ISM)}.\hskip 1em plus 0.5em minus 0.4em\relax IEEE, 2019, pp. 225--2255.

\bibitem{2018_IEEE_resunet}
Z.~Zhang, Q.~Liu, and Y.~Wang, ``Road extraction by deep residual {U-Net},''
  \emph{IEEE Geoscience and Remote Sensing Letters}, vol.~15, no.~5, pp.
  749--753, 2018.

\bibitem{2020_Springer_pranet}
D.-P. Fan, G.-P. Ji, T.~Zhou, G.~Chen, H.~Fu, J.~Shen, and L.~Shao, ``{PraNet}:
  Parallel reverse attention network for polyp segmentation,'' in
  \emph{International conference on medical image computing and
  computer-assisted intervention}.\hskip 1em plus 0.5em minus 0.4em\relax
  Springer, 2020, pp. 263--273.

\bibitem{2021_ArXiv_polypPVT}
B.~Dong, W.~Wang, D.-P. Fan, J.~Li, H.~Fu, and L.~Shao, ``{Polyp-PVT}: Polyp
  segmentation with pyramid vision transformers,'' \emph{arXiv preprint
  arXiv:2108.06932}, 2021.

\bibitem{vaswani2017attention}
A.~Vaswani, N.~Shazeer, N.~Parmar, J.~Uszkoreit, L.~Jones, A.~N. Gomez,
  {\L}.~Kaiser, and I.~Polosukhin, ``Attention is all you need,''
  \emph{Advances in neural information processing systems}, vol.~30, 2017.

\bibitem{2021_ACM_uacanet}
T.~Kim, H.~Lee, and D.~Kim, ``{UACANet}: Uncertainty augmented context
  attention for polyp segmentation,'' in \emph{Proceedings of the 29th ACM
  International Conference on Multimedia}, 2021, pp. 2167--2175.

\bibitem{2017_NeurIPS_meanTeacher}
A.~Tarvainen and H.~Valpola, ``Mean teachers are better role models:
  Weight-averaged consistency targets improve semi-supervised deep learning
  results,'' \emph{Advances in neural information processing systems}, vol.~30,
  2017.

\bibitem{2019_NeurIPS_mixmatch}
D.~Berthelot, N.~Carlini, I.~Goodfellow, N.~Papernot, A.~Oliver, and C.~A.
  Raffel, ``Mixmatch: A holistic approach to semi-supervised learning,''
  \emph{Advances in neural information processing systems}, vol.~32, 2019.

\bibitem{2022_Elseiver_ICT}
V.~Verma, K.~Kawaguchi, A.~Lamb, J.~Kannala, A.~Solin, Y.~Bengio, and
  D.~Lopez-Paz, ``Interpolation consistency training for semi-supervised
  learning,'' \emph{Neural Networks}, vol. 145, pp. 90--106, 2022.

\bibitem{2020_IJCNN_PL}
E.~Arazo, D.~Ortego, P.~Albert, N.~E. O’Connor, and K.~McGuinness,
  ``Pseudo-labeling and confirmation bias in deep semi-supervised learning,''
  in \emph{2020 International Joint Conference on Neural Networks
  (IJCNN)}.\hskip 1em plus 0.5em minus 0.4em\relax IEEE, 2020, pp. 1--8.

\bibitem{2019_CVPR_labelProp}
A.~Iscen, G.~Tolias, Y.~Avrithis, and O.~Chum, ``Label propagation for deep
  semi-supervised learning,'' in \emph{IEEE/CVF Conference on Computer Vision
  and Pattern Recognition (CVPR)}, 2019, pp. 5070--5079.

\bibitem{2021_AAAI_curriculumPL}
P.~Cascante-Bonilla, F.~Tan, Y.~Qi, and V.~Ordonez, ``Curriculum labeling:
  Revisiting pseudo-labeling for semi-supervised learning,'' in
  \emph{Proceedings of the AAAI Conference on Artificial Intelligence}, 2021.

\bibitem{2021_NeurIPS_flexmatch}
B.~Zhang, Y.~Wang, W.~Hou, H.~Wu, J.~Wang, M.~Okumura, and T.~Shinozaki,
  ``Flexmatch: Boosting semi-supervised learning with curriculum pseudo
  labeling,'' \emph{Advances in Neural Information Processing Systems},
  vol.~34, pp. 18\,408--18\,419, 2021.

\bibitem{2021_ICCV_PAWS}
M.~Assran, M.~Caron, I.~Misra, P.~Bojanowski, A.~Joulin, N.~Ballas, and
  M.~Rabbat, ``Semi-supervised learning of visual features by
  non-parametrically predicting view assignments with support samples,'' in
  \emph{Proceedings of the IEEE/CVF International Conference on Computer
  Vision}, 2021, pp. 8443--8452.

\bibitem{2018_ICLR_rotations}
S.~Gidaris, P.~Singh, and N.~Komodakis, ``Unsupervised representation learning
  by predicting image rotations,'' in \emph{International Conference on
  Learning Representations (ICLR)}, 2018.

\bibitem{2016_ECCV_colorization}
G.~Larsson, M.~Maire, and G.~Shakhnarovich, ``Learning representations for
  automatic colorization,'' in \emph{European conference on computer vision
  (ECCV)}.\hskip 1em plus 0.5em minus 0.4em\relax Springer, 2016, pp. 577--593.

\bibitem{2015_ICCV_context}
C.~Doersch, A.~Gupta, and A.~A. Efros, ``Unsupervised visual representation
  learning by context prediction,'' in \emph{International Conference on
  Computer Vision (ICCV)}, 2015, pp. 1422--1430.

\bibitem{2021_CVPR_simsiam}
X.~Chen and K.~He, ``Exploring simple siamese representation learning,'' in
  \emph{Conference on Computer Vision and Pattern Recognition (CVPR)}, 2021,
  pp. 15\,750--15\,758.

\bibitem{2021_ICLR_imix}
K.~Lee, Y.~Zhu, K.~Sohn, C.-L. Li, J.~Shin, and H.~Lee, ``{i-Mix}: A
  domain-agnostic strategy for contrastive representation learning,'' in
  \emph{International Conference on Learning Representations (ICLR)}, 2021.

\bibitem{2020_CVPR_MoCo}
K.~He, H.~Fan, Y.~Wu, S.~Xie, and R.~Girshick, ``{Momentum Contrast for
  Unsupervised Visual Representation Learning},'' in \emph{{Computer Vision and
  Pattern Recognition (CVPR)}}, 2020.

\bibitem{2021_ICCV_segmSSL}
J.~Yuan, Y.~Liu, C.~Shen, Z.~Wang, and H.~Li, ``A simple baseline for
  semi-supervised semantic segmentation with strong data augmentation,'' in
  \emph{Proceedings of the IEEE/CVF International Conference on Computer
  Vision}, 2021, pp. 8229--8238.

\bibitem{2022_CVPR_PLSegm}
Y.~Wang, H.~Wang, Y.~Shen, J.~Fei, W.~Li, G.~Jin, L.~Wu, R.~Zhao, and X.~Le,
  ``Semi-supervised semantic segmentation using unreliable pseudo-labels,'' in
  \emph{Proceedings of the IEEE/CVF Conference on Computer Vision and Pattern
  Recognition}, 2022, pp. 4248--4257.

\bibitem{2022_ICLR_bootstrapping}
S.~Liu, S.~Zhi, E.~Johns, and A.~J. Davison, ``Bootstrapping semantic
  segmentation with regional contrast,'' \emph{International Conference on
  Learning Representations (ICLR)}, 2021.

\bibitem{2019_Springer_medImgSSL}
H.~Shang, Z.~Sun, W.~Yang, X.~Fu, H.~Zheng, J.~Chang, and J.~Huang,
  ``Leveraging other datasets for medical imaging classification: evaluation of
  transfer, multi-task and semi-supervised learning,'' in \emph{International
  conference on medical image computing and computer-assisted
  intervention}.\hskip 1em plus 0.5em minus 0.4em\relax Springer, 2019, pp.
  431--439.

\bibitem{2020_CVPR_focalmix}
D.~Wang, Y.~Zhang, K.~Zhang, and L.~Wang, ``{FocalMix}: Semi-supervised
  learning for 3d medical image detection,'' in \emph{Proceedings of the
  IEEE/CVF Conference on Computer Vision and Pattern Recognition (CVPR)}, June
  2020.

\bibitem{2020_ArXiv_roam}
T.~Bdair, B.~Wiestler, N.~Navab, and S.~Albarqouni, ``{ROAM}: Random layer
  mixup for semi-supervised learning in medical imaging,'' \emph{arXiv preprint
  arXiv:2003.09439}, 2020.

\bibitem{2018_CVPR_learning}
S.~Sankaranarayanan, Y.~Balaji, A.~Jain, S.~N. Lim, and R.~Chellappa,
  ``Learning from synthetic data: Addressing domain shift for semantic
  segmentation,'' in \emph{Conference on computer vision and pattern
  recognition (CVPR)}, 2018.

\bibitem{2018_ECCV_synthUrban}
F.~S. Saleh, M.~S. Aliakbarian, M.~Salzmann, L.~Petersson, and J.~M. Alvarez,
  ``Effective use of synthetic data for urban scene semantic segmentation,'' in
  \emph{Proceedings of the European Conference on Computer Vision (ECCV)},
  September 2018.

\bibitem{2019_CVPR_SegmSynth}
Y.~Chen, W.~Li, X.~Chen, and L.~V. Gool, ``Learning semantic segmentation from
  synthetic data: A geometrically guided input-output adaptation approach,'' in
  \emph{Proceedings of the IEEE/CVF Conference on Computer Vision and Pattern
  Recognition (CVPR)}, June 2019.

\bibitem{2016_CVPR_synthia}
G.~Ros, L.~Sellart, J.~Materzynska, D.~Vazquez, and A.~M. Lopez, ``The
  {SYNTHIA} dataset: A large collection of synthetic images for semantic
  segmentation of urban scenes,'' in \emph{Proceedings of the IEEE Conference
  on Computer Vision and Pattern Recognition (CVPR)}, June 2016.

\bibitem{2018_ISBI_synthDAGAN}
M.~Frid-Adar, E.~Klang, M.~Amitai, J.~Goldberger, and H.~Greenspan, ``Synthetic
  data augmentation using {GAN} for improved liver lesion classification,'' in
  \emph{2018 IEEE 15th international symposium on biomedical imaging (ISBI
  2018)}.\hskip 1em plus 0.5em minus 0.4em\relax IEEE, 2018, pp. 289--293.

\bibitem{2021_ArXiv_improving}
D.~Schaudt, C.~Kloth, C.~Spaete, A.~Hinteregger, M.~Beer, and R.~von Schwerin,
  ``Improving {COVID-19} {CXR} detection with synthetic data augmentation,''
  \emph{arXiv preprint arXiv:2112.07529}, 2021.

\bibitem{2021_TMI_lowAvailability}
B.~Thamsen, P.~Yevtushenko, L.~Gundelwein, A.~A. Setio, H.~Lamecker, M.~Kelm,
  M.~Schafstedde, T.~Heimann, T.~Kuehne, and L.~Goubergrits, ``Synthetic
  database of aortic morphometry and hemodynamics: overcoming medical imaging
  data availability,'' \emph{IEEE Transactions on Medical Imaging}, vol.~40,
  no.~5, pp. 1438--1449, 2021.

\bibitem{2017_ArXiv_dualGAN}
J.~T. Guibas, T.~S. Virdi, and P.~S. Li, ``Synthetic medical images from dual
  generative adversarial networks,'' \emph{arXiv preprint arXiv:1709.01872},
  2017.

\bibitem{2014_NeurIPS_gans}
I.~Goodfellow, J.~Pouget-Abadie, M.~Mirza, B.~Xu, D.~Warde-Farley, S.~Ozair,
  A.~Courville, and Y.~Bengio, ``Generative adversarial nets,''
  \emph{{A}dvances in {N}eural {I}nformation {P}rocessing {S}ystems
  ({NeurIPS})}, 2014.

\bibitem{sanchez2020deep}
L.~F. Sanchez-Peralta, L.~Bote-Curiel, A.~Picon, F.~M. Sanchez-Margallo, and
  J.~B. Pagador, ``Deep learning to find colorectal polyps in colonoscopy: A
  systematic literature review,'' \emph{Artificial intelligence in medicine},
  vol. 108, p. 101923, 2020.

\bibitem{2018_ICLR_mixup}
H.~Zhang, M.~Cisse, Y.~Dauphin, and D.~Lopez-Paz, ``{mixup: Beyond Empirical
  Risk Minimization},'' in \emph{{International Conference on Learning
  Representations (ICLR)}}, 2018.

\bibitem{2020_TMI_threshold}
X.~Guo, C.~Yang, Y.~Liu, and Y.~Yuan, ``Learn to threshold: Thresholdnet with
  confidence-guided manifold mixup for polyp segmentation,'' \emph{IEEE
  transactions on medical imaging}, vol.~40, no.~4, pp. 1134--1146, 2020.

\bibitem{2019_ICCV_DeepVCP}
W.~Lu, G.~Wan, Y.~Zhou, X.~Fu, P.~Yuan, and S.~Song, ``Deepvcp: An end-to-end
  deep neural network for point cloud registration,'' in \emph{International
  Conference on Computer Vision (ICCV)}, October 2019.

\bibitem{2017_arxiv_deal}
M.~Lewis, D.~Yarats, Y.~N. Dauphin, D.~Parikh, and D.~Batra, ``Deal or no deal?
  end-to-end learning for negotiation dialogues,'' \emph{arXiv preprint
  arXiv:1706.05125}, 2017.

\bibitem{2021_CVPR_soho}
Z.~Huang, Z.~Zeng, Y.~Huang, B.~Liu, D.~Fu, and J.~Fu, ``Seeing out of the box:
  End-to-end pre-training for vision-language representation learning,'' in
  \emph{Conference on Computer Vision and Pattern Recognition (CVPR)}, June
  2021, pp. 12\,976--12\,985.

\bibitem{2017_JHE_CVC300}
D.~V{\'a}zquez, J.~Bernal, F.~J. S{\'a}nchez, G.~Fern{\'a}ndez-Esparrach, A.~M.
  L{\'o}pez, A.~Romero, M.~Drozdzal, and A.~Courville, ``A benchmark for
  endoluminal scene segmentation of colonoscopy images,'' \emph{Journal of
  healthcare engineering}, vol. 2017, 2017.

\bibitem{2015_Elseiver_CVCClinicDB}
J.~Bernal, F.~J. S{\'a}nchez, G.~Fern{\'a}ndez-Esparrach, D.~Gil,
  C.~Rodr{\'\i}guez, and F.~Vilari{\~n}o, ``{WM-DOVA} maps for accurate polyp
  highlighting in colonoscopy: Validation vs. saliency maps from physicians,''
  \emph{Computerized medical imaging and graphics}, vol.~43, pp. 99--111, 2015.

\bibitem{2012_PR_CVCColonDB}
J.~Bernal, J.~S{\'a}nchez, and F.~Vilarino, ``Towards automatic polyp detection
  with a polyp appearance model,'' \emph{Pattern Recognition}, vol.~45, no.~9,
  pp. 3166--3182, 2012.

\bibitem{2014_IJCARS_ETIS}
J.~Silva, A.~Histace, O.~Romain, X.~Dray, and B.~Granado, ``Toward embedded
  detection of polyps in {WCE} images for early diagnosis of colorectal
  cancer,'' \emph{International journal of computer assisted radiology and
  surgery}, vol.~9, no.~2, pp. 283--293, 2014.

\bibitem{2020_ICMM_kvasir}
D.~Jha, P.~H. Smedsrud, M.~A. Riegler, P.~Halvorsen, T.~d. Lange, D.~Johansen,
  and H.~D. Johansen, ``Kvasir-seg: A segmented polyp dataset,'' in
  \emph{International Conference on Multimedia Modeling}.\hskip 1em plus 0.5em
  minus 0.4em\relax Springer, 2020, pp. 451--462.

\bibitem{hung2018adversarial}
W.-C. Hung, Y.-H. Tsai, Y.-T. Liou, Y.-Y. Lin, and M.-H. Yang, ``Adversarial
  learning for semi-supervised semantic segmentation,'' \emph{arXiv preprint
  arXiv:1802.07934}, 2018.

\end{thebibliography}

\end{document}